\documentclass[11pt]{article}

\usepackage[preprint]{acl}

\usepackage{times}
\usepackage{latexsym}

\usepackage[T1]{fontenc}
\usepackage[utf8]{inputenc}

\usepackage{microtype}

\usepackage{inconsolata}

\usepackage{graphicx}
\usepackage{amsmath}
\usepackage{multirow} 
\usepackage{booktabs}
\usepackage{amssymb} 
\usepackage{bbm}

\usepackage{titlesec}
\titlespacing*{\section}      {0pt}{1.2ex plus .2ex minus .2ex}{0.8ex plus .1ex}
\titlespacing*{\subsection}   {0pt}{1.0ex plus .2ex minus .2ex}{0.6ex plus .1ex}
\titlespacing*{\subsubsection}{0pt}{0.8ex plus .1ex minus .1ex}{0.5ex plus .1ex}
\titlespacing*{\paragraph}{0pt}{0.5ex}{1em}
\usepackage{xcolor}
\usepackage{booktabs}
\usepackage{multirow}
\usepackage[table]{xcolor} 

\usepackage[T1]{fontenc}
\usepackage[most]{tcolorbox}
\usepackage{listings}

\lstdefinestyle{promptstyle}{
  basicstyle=\ttfamily\footnotesize,
  breaklines=true,
  breakatwhitespace=false,
  columns=fullflexible,
  keepspaces=true,
  showstringspaces=false,
  breakindent=0pt,
  postbreak=\mbox{\textcolor{gray}{$\hookrightarrow$}\space},
}

\newtcolorbox{preplanbox}[1]{
  colback=gray!5, colframe=gray!50,
  title=#1, fonttitle=\bfseries\small,
  fontupper=\small,
  left=3pt, right=3pt, top=3pt, bottom=3pt,
}

\lstdefinestyle{promptstyle}{
  basicstyle=\ttfamily\scriptsize,
  breaklines=true, breakatwhitespace=false,
  columns=fullflexible, keepspaces=true,
  showstringspaces=false, breakindent=0pt,
  postbreak=\mbox{\textcolor{gray}{$\hookrightarrow$}\space},
}
\newtcolorbox{promptbox}[1]{
  breakable, colback=gray!5, colframe=gray!50,
  title=#1, fonttitle=\bfseries,
  left=2pt, right=2pt, top=2pt, bottom=2pt,
}

\title{Knowing What to Solve Before How: Preplan Empowered LLM Mathematical Reasoning}

\author{
  \textbf{Shaojie Wang\textsuperscript{1}} \quad
  \textbf{Liang Zhang\textsuperscript{1}}\thanks{\ Corresponding author.} \\
  \\
  \texttt{\{shaojiewang, liangzhang\}@hkustgz.edu.cn} \\
  \\
  \textsuperscript{1} Hong Kong University of Science and Technology (Guangzhou) \\
}

\begin{document}
\maketitle
\begin{abstract}
Current plan-based reasoning methods improve large language models (LLMs) by inserting a planning stage before execution, giving rise to the \textbf{question $\rightarrow$ plan $\rightarrow$ cot} paradigm. While effective, a closer examination reveals an inherent paradigm-level gap: both the planning and its execution stages decide \textit{how to solve} a problem, while the prior question of \textit{what to solve}; recognizing the problem type, the applicable tools, and the foreseeable pitfalls; remains entirely implicit. To bridge this gap, we propose \textbf{PPC} (\textbf{P}replan-\textbf{P}lan-\textbf{C}oT), a framework that introduces an explicit problem-understanding stage, the \textit{preplan}, yielding a new \textbf{question $\rightarrow$ preplan $\rightarrow$ plan $\rightarrow$ cot} paradigm. Realizing this paradigm requires safeguarding the conceptual integrity of preplan at both ends. Specifically, we design a three-stage synthesis pipeline with a \textit{spoiler-score detector} that filters out leakage and spoiler failures to build clean preplan supervision, and a composite GRPO reward enforces that the generated plan genuinely follows from the preplan. Experiments across four backbones and five mathematical reasoning benchmarks show that PPC achieves the best results on 39 of 40 metrics, improving \emph{maj@16} and \emph{pass@16} by +2.23 and +3.06 over the strongest baseline without introducing additional inference token overhead.
% , demonstrating that modeling \textit{what to solve} before \textit{how to solve} improves accuracy without added generation cost.
\end{abstract}

\section{Introduction}
\label{sec:intro}

Large language models (LLMs) have made remarkable progress on mathematical reasoning tasks, driven by the Chain-of-Thought (CoT) paradigm~\citep{chain_of_thought,llm_zero_shot_reasoner}. Under this \textbf{question $\rightarrow$ cot} paradigm, the model produces, in a single forward pass, a long sequence that interleaves intermediate reasoning steps with the final answer. Combined with supervised fine-tuning (SFT) on CoT trajectories~\citep{mammoth} and reinforcement learning (RL) from verifiable rewards~\citep{deepseek_r1}, this paradigm has achieved strong results across a wide range of mathematical benchmarks~\citep{llm_for_math_reasoning}.

\noindent As problems grow more complex, however, single-pass CoT reveals a structural limitation: operating as an autoregressive, token-level process, it confines reasoning to \emph{step-level} execution and lacks higher-level organization, yielding redundant or incoherent trajectories. Recent work addresses this issue by inserting an intermediate planning stage, giving rise to the \textbf{question $\rightarrow$ plan $\rightarrow$ cot} paradigm: \emph{Plan-Tuning}~\citep{plantuning} distills planning trajectories from large models and fine-tunes smaller LLMs on (question, plan, solution) triples, while \emph{PTA-GRPO}~\citep{ptagrpo} introduces plan-level guidance rewards during reinforcement learning. Their shared insight is that an explicit \emph{plan-level} layer above \emph{step-level} execution supplies the global organization that single-pass CoT lacks.

\noindent However, a closer examination of the \textbf{question $\rightarrow$ plan $\rightarrow$ cot} paradigm reveals that its two stages, planning and execution, differ in granularity but share the same underlying role: both decide \emph{how to solve} the problem---the plan organizes the solution, and the execution carries it out. What neither stage addresses is the prior question of \emph{what to solve}: such as recognizing what type of problem it is, which tools or concepts may apply, which boundary conditions matter, and which pitfalls to anticipate. For example, given a problem $12x^2 - xy - 6y^2 = 0$, recognizing it as a homogeneous quadratic immediately signals the model to prioritize factorization over the more cumbersome discriminant method, while also indicating that each linear factor produces a family of solutions. Consequently, the subsequent plan can follow these hints and count solutions correctly. Existing methods implicitly assume that such understanding emerges for free during plan generation, leaving a \textbf{paradigm-level gap}: \emph{how to solve} is modeled explicitly, whereas \emph{what to solve} remains implicit. We refer to this missing stage as the \emph{preplan}: an explicit problem-understanding stage that precedes planning.

\begin{figure}[t]
  \centering
  \includegraphics[width=0.95\columnwidth]{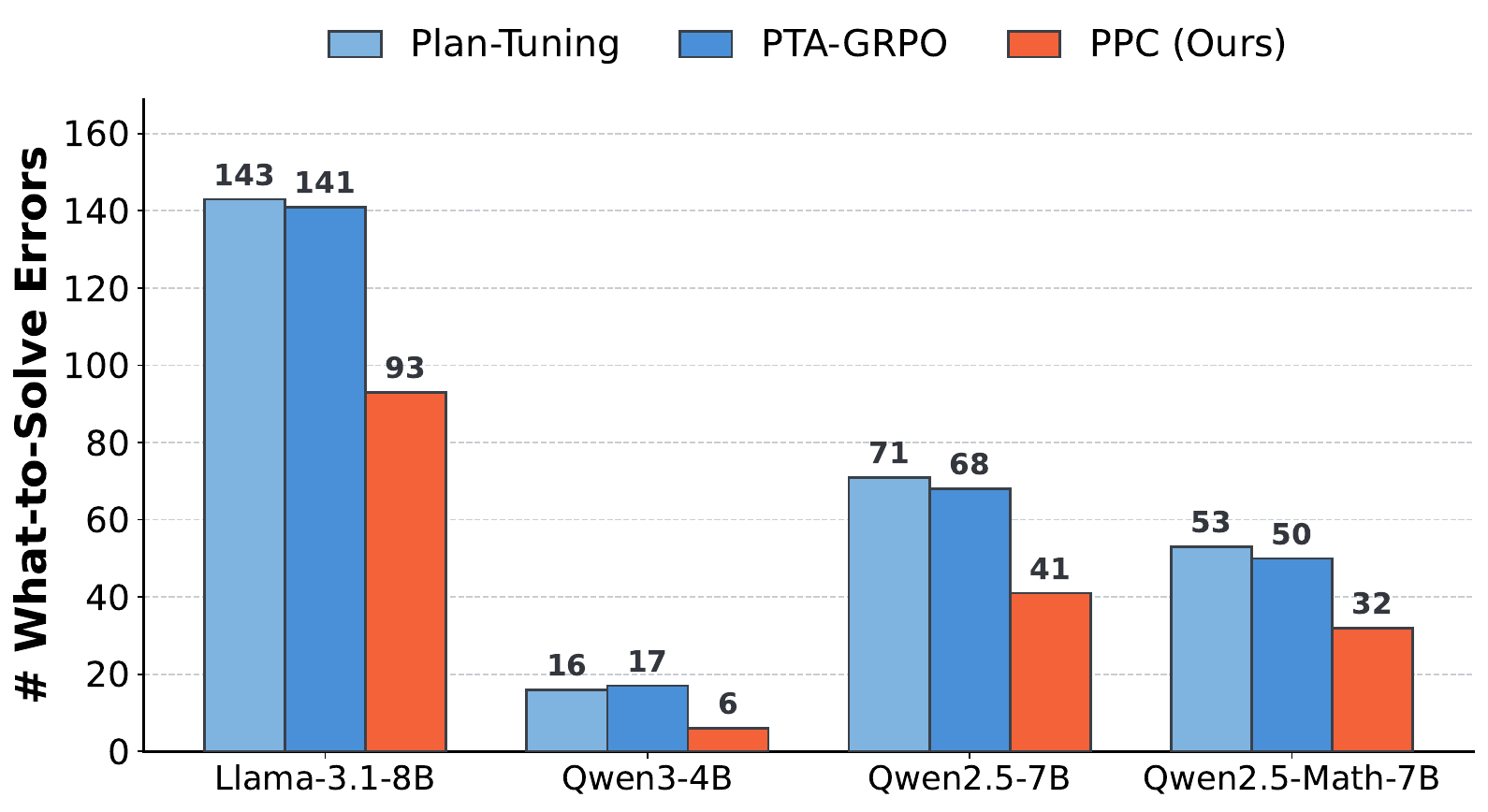}
  \caption{Number of \emph{what-to-solve} errors on MATH-500 across four backbones. Each wrong answer (under greedy decoding) is attributed to its root cause by an LLM judge, here we use DeepSeek-V4.}
  \label{fig:motivation}
  \vskip -0.2in
\end{figure}

\noindent Figure~\ref{fig:motivation} illustrates the cost of this paradigm-level gap. We diagnose the errors made by two representative \textbf{question $\rightarrow$ plan $\rightarrow$ cot} methods on MATH-500~\citep{math_500}, using an LLM judge to attribute each wrong answer to its root cause (see Appendix~\ref{appendix:analysis} for more details). Across four backbones, a large fraction of their errors stem not from calculation, but from a failure to understand \emph{what to solve}. This exposes the cost of leaving problem understanding implicit and motivates introducing it as an explicit stage prior to planning.

\noindent While conceptually intuitive, bridging this gap by directly introducing a problem-understanding stage faces two concrete challenges. \textbf{Challenge 1 (data construction).} No existing reasoning dataset provides supervision for such a preplan stage, and directly prompting a strong LLM to generate it tends to fail in two ways. The output either rehearses the forthcoming plan (\emph{leakage failure}), or quietly carries out intermediate computation under the guise of analysis (\emph{spoiler failure}). Both collapse the stage back into \emph{how to solve}, defeating its purpose. \textbf{Challenge 2 (training-time adherence).} Even given clean supervision, a model trained with standard outcome rewards may learn to emit a superficially well-formed analysis while ignoring it during plan generation, capturing the mere form of problem understanding without its practical utility. A successful framework must therefore enforce that the downstream plan genuinely follows from the preceding preplan.

\begin{figure}[t]
  \centering
  \includegraphics[width=0.95\columnwidth]{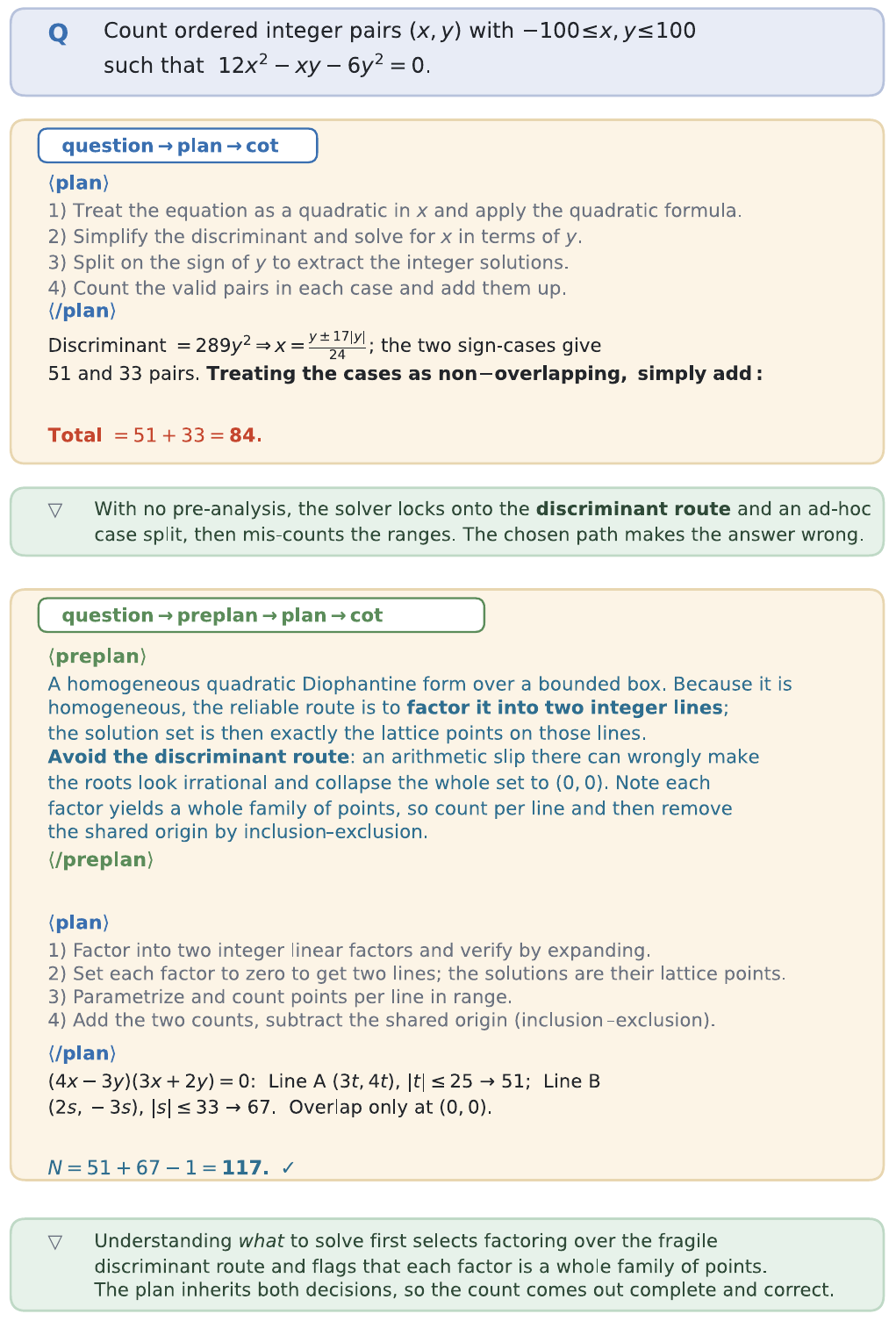}
  \caption{An example comparing the two paradigms on the same problem. \emph{Top}: a \textbf{question $\rightarrow$ plan $\rightarrow$ cot} solver commits to the discriminant route and undercounts ($84$). \emph{Bottom}: a preceding \emph{preplan} recognizes the homogeneous form, chooses factoring over the fragile discriminant route, and notes that each factor is a whole family of lattice points; the plan inherits this and the count is correct ($117$). Full trace in Appendix~\ref{app:spoiler_example}.}
  \label{fig:case}
  \vskip -0.15in
\end{figure}

\noindent To address these challenges, we propose \textbf{PPC} (\textbf{P}replan-\textbf{P}lan-\textbf{C}oT), a framework that explicitly introduces a problem-understanding stage, the \emph{preplan}, into both data construction and reinforcement learning, giving rise to the new \textbf{question $\rightarrow$ preplan $\rightarrow$ plan $\rightarrow$ cot} paradigm. A preplan is a comprehensive analysis about the problem that identifies the problem type, surveys relevant tools and concepts, flags the boundary conditions that matter, and anticipates pitfalls---it settles \emph{what to solve} and leaves \emph{how to solve} to the subsequent planning and execution stages. To address Challenge~1, we design a three-stage synthesis pipeline guarded by a \emph{spoiler-score detector} that filters out leakage and spoiler failures, yielding a clean reasoning dataset with explicit preplan supervision. For Challenge~2, we extend GRPO with a composite reward that jointly optimizes \emph{final-answer correctness} and \emph{plan-preplan alignment}, while enforcing structural compliance across the three stages and penalizing preplan style degradation through the same spoiler score.

\noindent Figure~\ref{fig:case} makes this new paradigm concrete: on the same problem, a \textbf{question $\rightarrow$ plan $\rightarrow$ cot} solver commits to a fragile route without analyzing what the problem requires, whereas an explicit preplan first recognizes the structure and selects a reliable route, guiding the subsequent plan toward the correct solution. This mirrors the aggregate trend in Figure~\ref{fig:motivation}. Our main contributions are as follows:

% A natural concern is that inserting an extra preplan stage inflates generation cost. PPC avoids this by design: rather than appending an auxiliary signal on top of the chain-of-thought, the preplan and plan jointly commit the model to a single solution path, so the execution advances along it instead of exploring redundant detours. The non-computational boundary further keeps the preplan itself short. 

\begin{itemize}
\item Conceptually: We identify and formalize a \textbf{paradigm-level gap} in existing plan-based reasoning: current methods model \emph{how to solve} explicitly but leave \emph{what to solve} implicit. To bridge it, we propose the \textbf{question $\rightarrow$ preplan $\rightarrow$ plan $\rightarrow$ cot} paradigm for the first time.

\item Methodologically: We construct the first reasoning dataset with explicit \emph{preplan} supervision, protected by a \emph{spoiler-score detector} against the two characteristic generation failures, leakage and spoiler. We then design a composite GRPO reward whose \emph{plan-preplan alignment} term explicitly enforces that the generated plan follows from the preceding preplan, preventing the model from generating superficially well-formed but otherwise ignored preplans.

\item Empirically: Across four backbones and five mathematical reasoning benchmarks, PPC achieves the best results on 39 of 40 metrics, improving \texttt{maj@16} and \texttt{pass@16} by +2.23 and +3.06 over the strongest baseline without introducing additional inference token overhead.
\end{itemize}

\section{Related Work}
\subsection{LLM Reasoning with Chain-of-Thought}
Chain-of-thought (CoT) prompting~\citep{chain_of_thought,llm_zero_shot_reasoner} established that eliciting intermediate reasoning steps before the final answer substantially improves LLM performance on complex reasoning tasks, and has since become the dominant paradigm for mathematical reasoning. To strengthen this capability beyond in-context prompting, two complementary post-training directions have been actively explored. The first relies on supervised fine-tuning over distilled CoT trajectories~\citep{openthoughts,limo,aimo-2}, where strong teacher models generate large pools of step-by-step solutions that are used to fine-tune smaller students. The second, popularized by DeepSeek-R1~\citep{deepseek_r1}, and subsequently extended by a growing line of work~\citep{deepseekmath,acereason,skywork_reasoner}, applies reinforcement learning with verifiable rewards(RLVR), in which models receive direct correctness feedback from deterministic verifiers and learn to produce long CoT trajectories with exploration. Despite these advances, recent analyses show that CoT trajectories 
generated by LLMs remain \emph{step-level}: the model derives each step locally as it decodes, without an explicit mechanism to commit to a global solution structure beforehand~\citep{ptagrpo, plan_solve_prompt}. This limitation motivates a complementary line of work that introduces higher-level structure into the reasoning process.
\subsection{Plan-Based Reasoning}
A complementary direction to scaling step-level CoT is to introduce \emph{higher-level structure} into the reasoning process. Early prompting-based work asks the model to plan before solving~\citep{plan_solve_prompt} or decomposes problems into simpler sub-questions~\citep{least-to-most}, with later extensions exploring hierarchical sub-goals~\citep{hypertree_planning, hierarchical_cot} and tree-structured exploration~\citep{tot}. They share the premise that organization above individual steps improves coherence on multi-stage problems. 
\noindent Recent work integrates planning into post-training. Plan-Tuning~\citep{plantuning} distills planning trajectories from large reasoning models and supervises smaller LLMs on (question, plan, solution) triples, while PTA-GRPO~\citep{ptagrpo} extends this to RL with a plan-aware reward that jointly evaluates plan quality and final answer correctness, elevating planning from an inference-time trick to a training-time objective. 
\noindent Yet all of these methods treat the \emph{plan} as the entry point of structured reasoning: both the planning and its execution answer \emph{how to solve the problem}, while the prior question of \emph{what} the problem is asking remains implicit. Our work targets this gap by introducing an explicit preplan stage upstream of planning, together with a training objective that enforces the downstream plan to genuinely follow from it.

\section{Methodology}
\label{sec:method} 
Our goal is to teach LLMs a new reasoning paradigm, \textbf{question $\rightarrow$ preplan $\rightarrow$ plan $\rightarrow$ cot}, in which a \emph{preplan} that analyzes \emph{what to solve} precedes the planning and execution that decide \emph{how to solve}. The defining property of the preplan is that it analyzes \emph{what to solve} rather than \emph{how}: it surveys problem type, tools, constraints, and pitfalls without committing to a derivation. This conceptual boundary is fragile at both ends of training. During \textbf{data construction}, a generator asked for a preplan tends to slip back into \emph{how to solve}, through leakage or spoiler; during \textbf{reinforcement learning}, a policy may reproduce the \emph{form} of a preplan while ignoring it when generating the following plan: the two failures of Challenge~1 and Challenge~2 (\S\ref{sec:intro}). 

\noindent PPC addresses both through a single organizing principle. Stage~1 (\S\ref{sec:sft}) supplies a \emph{clean demonstration} of the paradigm by filtering out leakage and spoiler, and Stage~2 (\S\ref{sec:rl}) enforces \emph{faithful use} through a composite reward. Crucially, the same rule-based signal $s(\cdot)$ patrols both ends: as a hard filter when building data (Eq.~\ref{eq:filter}) and as a soft penalty during RL, ensuring that the same conceptual integrity constraint is enforced.
\subsection{Preliminaries} \label{sec:prelim} 

\paragraph{Problem formulation.} Given a question $q$, the policy $\pi_\theta$ generates a structured trajectory $y = (y_{\text{pp}}, y_{\text{p}}, y_{\text{e}})$ enclosed by tags \texttt{<preplan>}, \texttt{<plan>}, \texttt{<execute>}, with the final answer $\hat{a}(y)$ parsed from a \texttt{\textbackslash boxed\{\}} marker in $y_{\text{e}}$. Unlike the prior \textbf{question$\rightarrow$\emph{plan}$\rightarrow$\emph{cot}} paradigm, we additionally require $y_{\text{pp}}$ to be a non-computational problem-understanding analysis upstream of the plan. 

\paragraph{GRPO.} We adopt Group Relative Policy Optimization~\citep{deepseekmath} as our RL backbone. For each prompt $q$, GRPO samples a group $\{y^{(i)}\}_{i=1}^{G} \sim \pi_{\theta_{\text{old}}}(\cdot \mid q)$ and computes group-normalized advantages 
\begin{equation} \hat{A}_i = \frac{R(y^{(i)}) - \text{mean}(\{R(y^{(j)})\}_{j=1}^{G})}{\text{std}(\{R(y^{(j)})\}_{j=1}^{G})} \label{eq:advantage} \end{equation} then optimizes the clipped policy objective with a KL regularizer to a reference policy $\pi_{\text{ref}}$: \begin{equation} \begin{split} \mathcal{J}_{\text{GRPO}}(\theta) = {E}_{q,\{y^{(i)}\}}\bigg[ & \frac{1}{G}\sum_{i=1}^{G}\min\!\big(\rho_i \hat{A}_i,\; \\ \text{clip}(\rho_i, 1\!-\!\epsilon, 1\!+\!\epsilon)\hat{A}_i\big) & - \beta\,{D}_{\text{KL}}\!\left[\pi_\theta \,\|\, \pi_{\text{ref}}\right]\bigg] 
\end{split}
\label{eq:grpo} 
\end{equation} 
where $\rho_i = \pi_\theta(y^{(i)} \mid q) / \pi_{\theta_{\text{old}}}(y^{(i)} \mid q)$ and $\epsilon, \beta$ are hyperparameters.

\subsection{Stage 1: Preplan-Supervised Data Construction}
\label{sec:sft}

The difficulty for dataset construction is not generation but \emph{conceptual integrity}. Specifically, a \emph{leakage} preplan typically rehearses the plan's step sequence; while a \emph{spoiler} preplan tends to substitute concrete computation for structural description. Both collapse the stage back into \emph{how to solve}. Our pipeline counters the first \emph{by construction} and the second \emph{by filtering}.

\paragraph{Three-stage synthesis.}
Each trajectory is generated by conditioning every stage only on its predecessors:
\begin{equation}
\begin{split}
y_{\text{pp}} \sim \pi_{\text{pp}}(\cdot \mid q), 
&\quad y_{\text{p}} \sim \pi_{\text{p}}(\cdot \mid q, y_{\text{pp}}),\\
\quad y_{\text{e}} \sim \pi_{\text{e}}(\cdot \mid q, y_{\text{p}})
\end{split}
\label{eq:synthesis}
\end{equation}
The strict left-to-right, stage-wise generation process allows finer-grained control over different components of the reasoning trajectory. Leakage is suppressed \emph{by construction} at the prompt level: $\pi_{\text{pp}}$ is prompted to forbid derivations, intermediate values, step-by-step procedures, and any reference to a solution path, so that the supervision targets \emph{what to solve} rather than \emph{how}. Implementation details are given in Appendix~\ref{app:three_stage_prompts}.

\paragraph{Spoiler-score filtering.}
Left-to-right ordering does not prevent \emph{spoiler}, since a generator can compute within a single stage. We therefore retain a trajectory only if its preplan is pure and its answer is correct:
\begin{equation}
s(y_{\text{pp}}) \le \tau_s
\quad\text{and}\quad
\hat{a}(y) \equiv a^\star
\label{eq:filter}
\end{equation}
where $s(\cdot) \in \{0,\dots,6\}$ is a rule-based \emph{spoiler score} that aggregates several derivation- and answer-revealing signals (defined in Appendix~\ref{app:spoiler_score}). The check is deliberately decoupled from correctness: a trajectory whose answer is correct is \emph{still discarded} if its preplan is impure. For instance, a preplan that already asserts computed classifications and rehearses the exact series-expansion procedure scores high and is filtered out despite reaching the correct answer (full example in Appendix~\ref{app:spoiler_example}). This separation is the point: the filter targets preplan purity, not answer correctness. The retained corpus $\mathcal{D}_{\text{SFT}}$ initializes the policy by token-level fine-tuning,
\begin{equation}
\mathcal{L}_{\text{SFT}}(\theta) = -\,{E}_{(q,y)\sim\mathcal{D}_{\text{SFT}}}
  \!\sum_{t} \log \pi_\theta(y_t \mid q, y_{<t})
\label{eq:sft}
\end{equation}
yielding the model $\pi_{\theta_0}$ that initializes the RL stage.

\subsection{Stage 2: Composite-Reward RL}
\label{sec:rl}

Clean demonstrations teach the policy to produce the paradigm's form, but imitation cannot guarantee the policy actually \emph{uses} each stage: a fluent preplan may be silently ignored when the plan is generated (Challenge~2). We therefore use reinforcement learning to anchor every stage to an explicit reward, replacing the binary outcome reward of standard GRPO with a four-term composite:
\begin{equation}
\begin{split}
R(y) = R_{\text{out}}(y) + &\lambda_a R_{\text{adh}}(y) + \lambda_f R_{\text{fmt}}(y) \\ -\lambda_s R_{\text{sty}}(y)
\end{split}
\label{eq:reward}
\end{equation}
with weights $\lambda_a, \lambda_f, \lambda_s > 0$. Intuitively, $R_{\text{out}}$ anchors \emph{still solving the problem}, $R_{\text{adh}}$ anchors \emph{the preplan being inherited by the plan}, and $R_{\text{fmt}}, R_{\text{sty}}$ prevent the first two from being gamed. All scores come from \emph{frozen} judges not updated during RL; exact mappings, weights, and judge prompts are in Appendix~\ref{app:reward_details}, with ranges summarized in Table~\ref{tab:reward}.

\paragraph{(1) Outcome reward with partial credit.}
A $0/1$ reward gives no gradient on near-miss trajectories whose preplan and plan are nonetheless informative. Following the partial-credit view of rubric-based rewards~\citep{rubrics}, we score incorrect answers by an LLM-judged proximity level $J_{\text{prox}}(y)$ rating \emph{solution-path quality} (not numerical closeness):
\begin{equation}
R_{\text{out}}(y) =
\begin{cases}
1, & \hat{a}(y) \equiv a^\star, \\[2pt]
g\big(J_{\text{prox}}(y)\big), & \text{otherwise}
\end{cases}
\label{eq:outcome}
\end{equation}
where $g$ is a monotone map capped strictly below $1$. The cap preserves a margin between near-miss and correct trajectories, thus the policy cannot inflate partial credit into the appearance of a correct answer.

\paragraph{(2) Plan--preplan adherence reward.}
An LLM critic $J_{\text{adh}}(\cdot,\cdot)$ scores the \emph{strategic alignment} between $y_{\text{pp}}$ and $y_{\text{p}}$, normalized to $R_{\text{adh}}(y) \in [0,1]$. The critic is instructed to judge alignment, \emph{not} plan quality; thus, a strong plan that ignores the preplan still obtains lower score. This also prevents the preplan from being decoratively produced but functionally ignored: tying the plan to the preplan rather than letting it become an independent solution.

\paragraph{(3) Structural format guard.}
$R_{\text{fmt}}(y) \in \{0, 1\}$ checks that the three tags each appear exactly once and in order and that a \texttt{\textbackslash boxed\{\}} marker lies inside $y_{\text{e}}$, preventing the other rewards from being satisfied by malformed outputs.

\paragraph{(4) Spoiler-style penalty.}
Reusing the spoiler score $s(\cdot)$ from Eq.~\eqref{eq:filter}, we apply a one-sided penalty $R_{\text{sty}}(y) = \max(0,\, s(y_{\text{pp}}) - \tau_s)$, activated only when the preplan drifts back toward derivation-heavy content during RL. Thus the conceptual integrity of preplan purified at construction time is held throughout training by the same signal.

\paragraph{Summary} 
$R_{\text{out}}$ keeps correctness as the dominant objective; $R_{\text{adh}}$ ensures the preplan is \emph{used} rather than \emph{produced}; $R_{\text{fmt}}$ and $R_{\text{sty}}$ prevent these signals from being satisfied through ill-formed outputs or by collapsing the preplan into computation. Notably, without $R_{\text{sty}}$ the model can satisfy adherence by writing a derivation-style preplan that trivially aligns with the downstream plan, defeating the purpose of our paradigm. We ablate each term in \S\ref{sec:ablation}.

\begin{table}[t]
\centering
\small
\caption{Reward components, their ranges, and roles. Correctness dominates;
adherence nudges; format and style act as guards.}
\label{tab:reward}
\begin{tabular}{@{}llp{2.7cm}@{}}
\toprule
\textbf{Term} & \textbf{Range} & \textbf{Role} \\
\midrule
$R_{\text{out}}$ & $[0,0.5]$ & main signal (partial credit) \\
$\lambda_a R_{\text{adh}}$ & $[0, \lambda_a]$ & adherence nudge \\
$\lambda_f R_{\text{fmt}}$ & $\{0, \lambda_f\}$ & structural guard \\
$-\lambda_s R_{\text{sty}}$ & $[-\,\lambda_s\!\cdot\!(6-\tau_s),\,0]$ & anti-degradation \\
\bottomrule
\end{tabular}
\end{table}

\section{Experiments}
\label{sec:exp}

\begin{table*}[t]
  \centering
  \small
  \setlength{\tabcolsep}{4pt}
  \setlength{\aboverulesep}{0pt}     
  \setlength{\belowrulesep}{0pt}      
  \renewcommand{\arraystretch}{1.25} 
  \newcommand{\best}[1]{\textbf{#1}}
  \newcommand{\snd}[1]{\underline{#1}}
  % 颜色定义
  \definecolor{ppcblue}{RGB}{205,224,245}  
  \definecolor{altgray}{RGB}{236,236,236}   
  \newcommand{\cb}{\cellcolor{ppcblue}}      
  \newcommand{\cg}{\cellcolor{altgray}}      
  \resizebox{\textwidth}{!}{%
  \begin{tabular}{ll cc cc cc cc cc}
    \toprule
    \multirow{2}{*}{\textbf{Model}} & \multirow{2}{*}{\textbf{Method}}
    & \multicolumn{2}{c}{\textbf{AIME25}}
    & \multicolumn{2}{c}{\textbf{Minerva-Math}}
    & \multicolumn{2}{c}{\textbf{OlympiadBench}}
    & \multicolumn{2}{c}{\textbf{MATH-500}}
    & \multicolumn{2}{c}{\textbf{GSM8K}} \\
    \cmidrule(lr){3-4}\cmidrule(lr){5-6}\cmidrule(lr){7-8}\cmidrule(lr){9-10}\cmidrule(lr){11-12}
    & & maj@16 & pass@16 & maj@16 & pass@16 & maj@16 & pass@16 & maj@16 & pass@16 & maj@16 & pass@16 \\
    \midrule
    \multirow{6}{*}{Qwen3-4B}
    & Base        & 60.00 & 73.33 & 43.01 & 55.15 & 66.04 & 76.81 & 96.00 & 98.20 & 94.84 & 96.89 \\
    & \cg Prompt-Only & \cg 60.00 & \cg 76.67 & \cg 43.75 & \cg 54.41 & \cg 66.48 & \cg 77.25 & \cg 96.40 & \cg 98.20 & \cg 94.62 & \cg 97.27 \\
    & GRPO        & 60.00 & \snd{76.67} & 44.48 & 55.88 & 64.62 & \snd{77.91} & \snd{96.60} & \snd{98.80} & 94.92 & 97.04 \\
    & \cg PTA-GRPO & \cg 60.00 & \cg \snd{76.67} & \cg \snd{45.22} & \cg \snd{58.09} & \cg 59.89 & \cg 76.15 & \cg 95.80 & \cg 98.60 & \cg \best{95.30} & \cg \snd{97.57} \\
    & Plan-Tuning & 56.67 & \snd{76.67} & 44.85 & 56.25 & 66.04 & 77.25 & 96.20 & \snd{98.80} & 95.00 & 97.12 \\
    & \cb \textbf{PPC} & \cb \best{63.33} & \cb \best{80.00} & \cb \best{46.32} & \cb \best{60.66} & \cb \best{67.03} & \cb \best{78.35} & \cb \best{97.20} & \cb \best{99.40} & \cb \snd{95.15} & \cb \best{98.03} \\
    \midrule
    \multirow{6}{*}{Qwen2.5-7B}
    & Base        & 10.00 & 30.00 & 40.07 & 54.78 & 40.11 & 57.47 & 80.60 & 89.00 & 93.71 & 97.04 \\
    & \cg Prompt-Only & \cg 16.67 & \cg 30.00 & \cg 39.71 & \cg 58.09 & \cg 40.77 & \cg 61.32 & \cg 80.40 & \cg 90.00 & \cg 94.47 & \cg 97.12 \\
    & GRPO        & \snd{16.67} & \snd{30.00} & 41.18 & \snd{61.76} & \snd{43.19} & 59.56 & 83.80 & \snd{90.80} & 94.31 & \snd{97.57} \\
    & \cg PTA-GRPO & \cg \snd{16.67} & \cg 23.33 & \cg \snd{43.01} & \cg 58.82 & \cg 40.77 & \cg \snd{62.09} & \cg 81.40 & \cg 89.40 & \cg \snd{94.62} & \cg 97.27 \\
    & Plan-Tuning & \best{20.00} & 26.67 & 41.54 & 55.88 & 41.53 & 58.13 & \snd{84.00} & 88.80 & 94.09 & 97.19 \\
    & \cb \textbf{PPC} & \cb \best{20.00} & \cb \best{36.67} & \cb \best{44.85} & \cb \best{62.87} & \cb \best{44.73} & \cb \best{63.85} & \cb \best{84.80} & \cb \best{92.20} & \cb \best{95.22} & \cb \best{98.48} \\
    \midrule
    \multirow{6}{*}{Qwen2.5-Math-7B}
    & Base        & \snd{20.00} & 30.00 & 40.81 & 51.84 & 43.41 & 53.96 & 87.00 & 90.80 & \snd{96.97} & 98.10 \\
    & \cg Prompt-Only & \cg 16.67 & \cg 30.00 & \cg 41.92 & \cg 52.21 & \cg 44.62 & \cg 64.51 & \cg 87.60 & \cg 91.80 & \cg 97.04 & \cg 98.18 \\
    & GRPO        & \snd{20.00} & \snd{33.33} & \snd{42.65} & 53.31 & \snd{45.05} & 64.18 & 86.80 & 93.40 & 96.36 & \snd{98.40} \\
    & \cg PTA-GRPO & \cg 16.67 & \cg 23.33 & \cg 37.87 & \cg \snd{56.62} & \cg 43.52 & \cg 61.54 & \cg \snd{87.80} & \cg 92.80 & \cg 95.15 & \cg 98.03 \\
    & Plan-Tuning & 16.67 & 26.67 & 40.07 & 52.94 & 43.95 & \snd{64.73} & 87.60 & \snd{93.60} & 96.82 & 98.33 \\
    & \cb \textbf{PPC} & \cb \best{23.33} & \cb \best{36.67} & \cb \best{43.38} & \cb \best{58.09} & \cb \best{46.04} & \cb \best{66.26} & \cb \best{88.60} & \cb \best{95.00} & \cb \best{97.12} & \cb \best{98.56} \\
    \midrule
    \multirow{6}{*}{Llama3.1-8B}
    & Base        & 0.00 & 3.33 & 35.29 & \snd{60.29} & 21.10 & 41.32 & 55.20 & 82.00 & 89.76 & 95.98 \\
    & \cg Prompt-Only & \cg 0.00 & \cg 3.33 & \cg 34.56 & \cg 59.56 & \cg 23.08 & \cg 42.42 & \cg 58.00 & \cg 80.00 & \cg 90.37 & \cg 96.47 \\
    & GRPO        & 0.00 & 10.00 & 36.40 & 59.19 & \snd{25.05} & 42.20 & 59.60 & \snd{83.40} & 90.59 & 96.36 \\
    & \cg PTA-GRPO & \cg \best{6.67} & \cg 10.00 & \cg \snd{37.13} & \cg 59.93 & \cg 24.40 & \cg \snd{47.47} & \cg \snd{60.60} & \cg 82.80 & \cg 90.75 & \cg \snd{96.59} \\
    & Plan-Tuning & 0.00 & \snd{13.33} & 36.03 & 58.09 & 22.64 & 47.14 & 60.20 & 83.00 & \snd{90.83} & 96.13 \\
    & \cb \textbf{PPC} & \cb \best{6.67} & \cb \best{20.00} & \cb \best{37.50} & \cb \best{61.40} & \cb \best{27.80} & \cb \best{51.43} & \cb \best{65.80} & \cb \best{85.00} & \cb \best{92.04} & \cb \best{98.03} \\
    \bottomrule
  \end{tabular}%
  }
  \caption{\label{tab:main-results}
    Main results on five mathematical reasoning benchmarks across four backbones. We report maj@16 (self-consistency majority vote) and pass@16 for each benchmark. Within each backbone group and column, \textbf{bold} marks the best result and \underline{underline} the second best. Rows of our method (\textbf{PPC}) are highlighted in light blue.}
\end{table*}

\subsection{Experimental Setup}
\label{sec:setup}

\paragraph{Training data.}
We build our training corpus from DeepMath-103K~\citep{deepmath}, a large-scale mathematical reasoning dataset with per-problem difficulty annotations. We retain problems spanning four difficulty levels from medium to competition-hard, and stratify-sample a non-overlapping SFT/RL split of problems. SFT trajectories are then synthesized and filtered as described in \S\ref{sec:sft}, yielding the final training set. 

\paragraph{Models.}
To evaluate generality across model scales and families, we apply our framework to four base models: Qwen2.5-7B-Instruct~\citep{qwen2}, Qwen2.5-Math-7B-Instruct~\citep{qwen2.5_math}, LLaMA-3.1-8B-Instruct~\citep{llama3.1} and Qwen3-4B-Instruct~\citep{qwen3}. 

\paragraph{Benchmarks.}
We evaluate on five mathematical reasoning benchmarks spanning a wide difficulty range: AIME25, Minerva-Math~\citep{minerva-math}, OlympiadBench~\citep{olympiadbench}, MATH-500~\citep{math_500} and GSM8K~\citep{gsm8k}.

\paragraph{Baselines.}
We compare against the following methods, all trained on the same data splits for fairness:
(i) \textbf{Base}: the untrained backbone with zero-shot CoT prompting;
(ii) \textbf{Prompt-Only}: the same untrained backbone prompted to first analyze the problem itself:its type, relevant concepts, strategic direction, constraints, and pitfalls; and then form a plan before solving; this elicits PPC's preplan-then-solve structure purely through instruction, isolating whether prompt engineering alone suffices and thereby testing the necessity of our training;
(iii) \textbf{GRPO}~\citep{deepseekmath}: standard GRPO with a binary outcome reward, initialized from vanilla SFT;
(iv) \textbf{Plan-Tuning}~\citep{plantuning}: a post-training framework that distills planning trajectories from larger LLMs and fine-tunes the model via combined supervised and reinforcement-learning objectives;
(v) \textbf{PTA-GRPO}~\citep{ptagrpo}: a two-stage framework that first performs SFT on distilled high-level guidance and then applies GRPO with a plan-aware reward.

\paragraph{Evaluation protocol.} 
For each problem we sample $k=16$ trajectories and report \emph{maj@16} (self-consistency majority vote) as our primary metric, together with \emph{pass@16} as an upper-bound coverage measure. All methods use the identical chat template and system prompt (``\emph{Please reason step by step, and put your final answer within \textbackslash boxed\{\}}''. Final answers are extracted from the last \texttt{\textbackslash boxed\{\}} marker and scored by the same answer-equivalence verifier used in training (\S\ref{sec:rl}).

\paragraph{Implementation details.}
\textbf{Data synthesis:} The three-stage trajectories of \S\ref{sec:sft} are generated with Qwen3-235B~\citep{qwen3} as the preplan and plan generators ($\pi_{\text{pp}}, \pi_{\text{p}}$) and DeepSeek-R1~\citep{deepseek_r1} as the execution model ($\pi_{\text{e}}$); the raw solution is then reorganized into a numbered execution by the same instruction-tuned model. The spoiler filter (Eq.~\ref{eq:filter}) uses threshold $\tau_s = 2$ and retains preplans of length $\ell_{\min}=150$ to $\ell_{\max}=1500$ tokens.

\noindent \textbf{Training:} We initialize from each backbone via SFT on $\mathcal{D}_{\text{SFT}}$ for $3$ epochs with learning rate $1e-5$ and batch size $16$, then run composite-reward GRPO for 500 steps. GRPO uses group size $G=8$ and reward weights in Eq.~\eqref{eq:reward} are $\lambda_a=0.1$, $\lambda_f=0.3$, and $\lambda_s=0.1$.

\noindent \textbf{Generation.} For both training rollouts and evaluation we sample with temperature $T=1.0$ and top-$p=0.95$. All experiments are run on 4 $\times$ NVIDIA RTX PRO 6000 Blackwell (96GB) GPUs.

\begin{table*}[t]
  \centering
  \small
  \setlength{\tabcolsep}{4pt}
  \setlength{\aboverulesep}{0pt}     
  \setlength{\belowrulesep}{0pt}      
  \newcommand{\best}[1]{\textbf{#1}}          
  \definecolor{ppcblue}{RGB}{205,224,245}       
  \definecolor{altgray}{RGB}{236,236,236}       
  \newcommand{\cb}{\cellcolor{ppcblue}}         
  \newcommand{\cg}{\cellcolor{altgray}}          
  \resizebox{\textwidth}{!}{%
  \begin{tabular}{ll cc cc cc cc cc}
    \toprule
    \multirow{2}{*}{\textbf{Model}} & \multirow{2}{*}{\textbf{Method}}
    & \multicolumn{2}{c}{\textbf{AIME25}}
    & \multicolumn{2}{c}{\textbf{Minerva-Math}}
    & \multicolumn{2}{c}{\textbf{OlympiadBench}}
    & \multicolumn{2}{c}{\textbf{MATH-500}}
    & \multicolumn{2}{c}{\textbf{GSM8K}} \\
    \cmidrule(lr){3-4}\cmidrule(lr){5-6}\cmidrule(lr){7-8}\cmidrule(lr){9-10}\cmidrule(lr){11-12}
    & & maj@16 & pass@16 & maj@16 & pass@16 & maj@16 & pass@16 & maj@16 & pass@16 & maj@16 & pass@16 \\
    \midrule
    \multirow{4}{*}{Qwen3-4B}
    & $R_{\text{out}}$ only            & 46.67 & 66.67 & 44.85 & 57.35 & 64.40 & 75.05 & 93.40 & 98.20 & 94.01 & 97.27 \\
    & \cg \quad$+\,R_{\text{sty}}$     & \cg 53.33 & \cg 70.00 & \cg 45.22 & \cg 58.46 & \cg 65.16 & \cg 77.58 & \cg 95.60 & \cg 98.60 & \cg 94.54 & \cg 97.34 \\
    & \quad$+\,R_{\text{adh}}$         & 60.00 & 76.67 & 45.96 & 59.56 & 66.37 & 78.02 & 96.60 & 98.80 & 94.92 & 97.80 \\
    & \cb \textbf{PPC} (full)          & \cb \best{63.33} & \cb \best{80.00} & \cb \best{46.32} & \cb \best{60.66} & \cb \best{67.03} & \cb \best{78.35} & \cb \best{97.20} & \cb \best{99.40} & \cb \best{95.15} & \cb \best{98.03} \\
    \midrule
    \multirow{4}{*}{Qwen2.5-7B}
    & $R_{\text{out}}$ only            & 10.00 & 23.33 & 42.28 & 58.82 & 41.98 & 62.20 & 82.20 & 91.00 & 94.39 & 97.57 \\
    & \cg \quad$+\,R_{\text{sty}}$     & \cg 13.33 & \cg 26.67 & \cg 42.87 & \cg 60.29 & \cg 43.38 & \cg 62.96 & \cg 83.40 & \cg 91.60 & \cg 94.47 & \cg 97.95 \\
    & \quad$+\,R_{\text{adh}}$         & 16.67 & 30.00 & 44.12 & 61.03 & 43.40 & 63.07 & 84.00 & 91.80 & 94.84 & 98.10 \\
    & \cb \textbf{PPC} (full)          & \cb \best{20.00} & \cb \best{36.67} & \cb \best{44.85} & \cb \best{62.87} & \cb \best{44.73} & \cb \best{63.85} & \cb \best{84.80} & \cb \best{92.20} & \cb \best{95.22} & \cb \best{98.48} \\
    \midrule
    \multirow{4}{*}{Qwen2.5-Math-7B}
    & $R_{\text{out}}$ only            & 16.67 & 23.33 & 41.18 & 56.25 & 42.42 & 63.41 & 86.40 & 93.40 & 96.74 & 98.10 \\
    & \cg \quad$+\,R_{\text{sty}}$     & \cg 20.00 & \cg 26.67 & \cg 41.91 & \cg 56.62 & \cg 43.08 & \cg 64.39 & \cg 87.40 & \cg 94.20 & \cg 96.89 & \cg 98.18 \\
    & \quad$+\,R_{\text{adh}}$         & 20.00 & 33.33 & 42.27 & 57.35 & 44.29 & 65.05 & 87.80 & 94.60 & 96.97 & 98.33 \\
    & \cb \textbf{PPC} (full)          & \cb \best{23.33} & \cb \best{36.67} & \cb \best{43.38} & \cb \best{58.09} & \cb \best{46.04} & \cb \best{66.26} & \cb \best{88.60} & \cb \best{95.00} & \cb \best{97.12} & \cb \best{98.56} \\
    \midrule
    \multirow{4}{*}{Llama3.1-8B}
    & $R_{\text{out}}$ only            & 0.00 & 10.00 & 35.66 & 58.82 & 23.74 & 48.35 & 63.80 & 82.80 & 90.98 & 97.12 \\
    & \cg \quad$+\,R_{\text{sty}}$     & \cg 3.33 & \cg 10.00 & \cg 36.03 & \cg 60.29 & \cg 25.27 & \cg 49.23 & \cg 64.60 & \cg 83.60 & \cg 91.35 & \cg 97.19 \\
    & \quad$+\,R_{\text{adh}}$         & 6.67 & 16.67 & 36.76 & 61.03 & 26.37 & 51.10 & 65.00 & 84.80 & 91.73 & 97.42 \\
    & \cb \textbf{PPC} (full)          & \cb \best{6.67} & \cb \best{20.00} & \cb \best{37.50} & \cb \best{61.40} & \cb \best{27.80} & \cb \best{51.43} & \cb \best{65.80} & \cb \best{85.00} & \cb \best{92.04} & \cb \best{98.03} \\
    \bottomrule
  \end{tabular}%
  }
  \caption{\label{tab:ablation}
    Ablation over reward components, starting from $R_{\text{out}}$ and cumulatively adding $R_{\text{sty}}$, $R_{\text{adh}}$. $R_{\text{fmt}}$ is kept across all settings. Each row adds one term to the row above. We report maj@16 and pass@16; the best result within each backbone group is in \textbf{bold}, and our full method (\textbf{PPC}) rows are highlighted in light blue.}
\end{table*}

\subsection{Experimental Results}
We organize our evaluation around four questions: whether PPC improves accuracy over plan-based baselines (\S\ref{sec:main_results}), whether each reward component contributes (\S\ref{sec:ablation}), whether the extra preplan stage inflates inference cost (\S\ref{sec:token}), and whether the preplan is genuinely used rather than merely produced (\S\ref{sec:faithful}). 
\subsubsection{Main Results}
\label{sec:main_results}
We evaluate PPC against five baselines: Base, Prompt-Only, GRPO, PTA-GRPO, and Plan-Tuning on five mathematical reasoning benchmarks (AIME25, Minerva-Math, OlympiadBench, MATH-500, GSM8K) across four backbones spanning different families and scales, reporting both maj@16 and pass@16 (Table~\ref{tab:main-results}).

\noindent PPC achieves the best results on 39 of 40 metrics, with consistent gains across all backbones. The improvements are largest on competition-level benchmarks: on Llama3.1-8B, PPC raises MATH-500 maj@16 by $+10.6$ over Base and OlympiadBench pass@16 from $41.3$ to $51.4$, while on the near-saturated GSM8K, all methods cluster within a narrow band. The Prompt-Only baseline, which prompts the untrained backbone to first analyze \emph{what to solve} before planning and executing, yields marginal and inconsistent improvements over Base: it improves some settings but trails plain GRPO on the large majority of metrics and never approaches PPC, indicating that \emph{prompting} for the preplan cannot substitute for \emph{learning} it. Notably, the prior plan-based methods do not consistently beat plain GRPO on several Minerva-Math and OlympiadBench settings, and occasionally fall below Base. Since PPC differs from these baselines by the upstream preplan, its consistent advantage isolates the effect of modeling \emph{what to solve} before \emph{how to solve}, confirming the paradigm-level gap identified in our paper.
\subsubsection{Ablation Study}
\label{sec:ablation}

We ablate the reward components by starting from the outcome reward $R_{\text{out}}$ alone and cumulatively adding the spoiler-style penalty $R_{\text{sty}}$, the adherence reward $R_{\text{adh}}$,  evaluating each configuration on all five benchmarks across four backbones (Table~\ref{tab:ablation}). The format guard $R_{\text{fmt}}$ is kept across all settings.  

\noindent Every term improves performance monotonically across all four backbones, indicating that the four terms are complementary rather than redundant. Adding $R_{\text{sty}}$ alone already lifts AIME25 maj@16 by $+6.7$ on Qwen3-4B, showing that keeping the preplan conceptual integrity is beneficial in itself and must be actively maintained during RL, not merely inherited from initialization. The adherence reward contributes its largest increments: a further $+6.7$ AIME25 maj@16 on Qwen3-4B and $+1.9$ OlympiadBench pass@16 on Llama3.1-8B: while its effect on the near-saturated GSM8K is marginal, supporting our claim that the preplan pays off only when the plan genuinely follows it. The gap between $R_{\text{out}}$ and full PPC widens on competition-level tasks (up to $+16.7$ AIME25 maj@16 on Qwen3-4B) and shrinks on GSM8K, mirroring the main-results trend that explicit problem understanding matters most when the solution path is hard to find.

\begin{table*}[t]
  \centering
  \small
  \setlength{\tabcolsep}{4pt}
  \setlength{\aboverulesep}{0pt}     
  \setlength{\belowrulesep}{0pt}    
  \newcommand{\best}[1]{\textbf{#1}}           
  \newcommand{\snd}[1]{\underline{#1}}         
  \definecolor{ppcblue}{RGB}{205,224,245}      
  \definecolor{altgray}{RGB}{236,236,236}       
  \newcommand{\cb}{\cellcolor{ppcblue}}         
  \newcommand{\cg}{\cellcolor{altgray}}        
  \resizebox{0.85\textwidth}{!}{%
  \begin{tabular}{l cc cc cc cc}
    \toprule
    \multirow{2}{*}{\textbf{Setting}}
    & \multicolumn{2}{c}{\textbf{Qwen3-4B}}
    & \multicolumn{2}{c}{\textbf{Qwen2.5-7B}}
    & \multicolumn{2}{c}{\textbf{Qwen2.5-Math-7B}}
    & \multicolumn{2}{c}{\textbf{Llama3.1-8B}} \\
    \cmidrule(lr){2-3}\cmidrule(lr){4-5}\cmidrule(lr){6-7}\cmidrule(lr){8-9}
    & maj@16 & pass@16 & maj@16 & pass@16 & maj@16 & pass@16 & maj@16 & pass@16 \\
    \midrule
    Shuffled       & \snd{96.40} & \snd{98.40} & \snd{82.60} & \snd{91.40} & \snd{87.20} & \snd{94.00} & 57.80 & 80.80 \\
    \cg Mismatched & \cg 46.80 & \cg 79.20 & \cg 27.80 & \cg 48.60 & \cg 36.40 & \cg 62.80 & \cg 37.00 & \cg 67.80 \\
    Generic        & 94.20 & 96.80 & 80.20 & 90.60 & 86.40 & 93.60 & \snd{59.40} & \snd{82.00} \\
    \cb \textbf{PPC} & \cb \best{97.20} & \cb \best{99.40} & \cb \best{84.80} & \cb \best{92.20} & \cb \best{88.60} & \cb \best{95.00} & \cb \best{65.80} & \cb \best{85.00} \\
    \bottomrule
  \end{tabular}%
  }
  \caption{\label{tab:faithfulness}
    Faithfulness analysis on MATH-500. We compare PPC against three preplan-perturbation baselines: \emph{Shuffled} (reordered preplan steps), \emph{Mismatched} (a preplan from a different problem), and \emph{Generic} (a content-free generic preplan). We report maj@16 and pass@16; within each column the best result is in \textbf{bold} and the second best is \underline{underlined}. Our method (\textbf{PPC}) rows are highlighted in light blue.}
\end{table*}

\begin{table}[t]
\centering
% \small
\setlength{\tabcolsep}{5pt}
\setlength{\aboverulesep}{0pt}
\setlength{\belowrulesep}{0pt}
\newcommand{\best}[1]{\textcolor{red}{#1}}
\definecolor{ppcblue}{RGB}{205,224,245}
\definecolor{altgray}{RGB}{236,236,236}
\newcommand{\cg}{\cellcolor{altgray}}
\caption{Average generated tokens per problem on MATH-500 (in K tokens). Despite the extra preplan stage, PPC is more compact than full CoT on every backbone and far more compact than PTA-GRPO.}
\label{tab:token}
\resizebox{\columnwidth}{!}{%
\begin{tabular}{@{}lccc >{\columncolor{ppcblue}[\tabcolsep][0pt]}c @{}}
\toprule
\textbf{Backbone} & \textbf{GRPO} & \textbf{Plan-Tuning} & \textbf{PTA-GRPO} & \textbf{PPC} \\
\midrule
Llama3.1          & 2.67 & 2.55 & 5.85 & 2.41 \\
\cg Qwen3         & \cg 1.80 & \cg 1.63 & \cg 4.18 & 1.60 \\
Qwen2.5           & 3.02 & 1.44 & 9.97 & 1.47 \\
\cg Qwen2.5-Math  & \cg 1.08 & \cg 0.96 & \cg 1.16 & 1.02 \\
\bottomrule
\end{tabular}
}
\end{table}

\subsubsection{Inference Efficiency} 
\label{sec:token} 
To test whether inserting an extra preplan stage inflates inference cost, we measure the average number of generated tokens per problem on MATH-500 for GRPO, Plan-Tuning, PTA-GRPO, and PPC across four backbones (Table~\ref{tab:token}). 

\noindent Despite producing an additional preplan, PPC generates fewer tokens than full CoT on all four backbones (e.g., $1.47$k vs.\ $3.02$k on Qwen2.5-7B) and far fewer than PTA-GRPO, whose unconstrained reasoning grows to $9.97$k tokens on the same backbone; PPC remains within a narrow band of the most compact baseline, Plan-Tuning, while achieving uniformly higher accuracy (\S\ref{sec:main_results}). The reason is that the preplan and plan jointly constrain execution rather than lengthen it: a single-pass CoT explores without higher-level structure and produces redundant detours, whereas in PPC the execution follows a committed plan and advances along a single solution path. This also explains PTA-GRPO's inflated length: it adds a plan as an auxiliary guidance signal on top of the CoT without constraining the reasoning to follow it, therefore, the plan lengthens the output without pruning the underlying exploration. The tokens saved by avoiding wrong-path detours outweigh the small fixed overhead of the preplan, thus PPC improves accuracy and reduces token cost simultaneously rather than trading one for the other.

\subsubsection{Faithfulness} 
\label{sec:faithful} 
To test whether the preplan is genuinely used rather than just produced, we intervene on it at inference time and measure the downstream effects. We replace the model-generated preplan with three perturbations while letting the model freely generate the plan and execution from it: \emph{Shuffled} (the same preplan with its sentences reordered), \emph{Mismatched} (a preplan taken from a different problem), and \emph{Generic} (a content-free placeholder). We report maj@16 and pass@16 on MATH-500 across four backbones (Table~\ref{tab:faithfulness}). 

\noindent Injecting a mismatched preplan causes severe collapse: maj@16 drops from $97.2$ to $46.8$ on Qwen3-4B and from $84.8$ to $27.8$ on Qwen2.5-7B. This shows that the preplan is genuinely used, not merely produced: if it were ignored, an unrelated preplan would be harmless and not mislead the downstream planning and execution.

\noindent The Shuffled and Generic conditions jointly confirm that this dependence is on the preplan's \emph{information}, not an artifact of the intervention. A generic, content-free preplan causes a marginal drop in accuracy ($94.2$ vs.\ $97.2$ on Qwen3-4B, $80.2$ vs.\ $84.8$ on Qwen2.5-7B), serving as a placebo: it confirms that the collapse under Mismatched stems from the injected content, not from the intervention itself. Against this baseline, the contrast with Shuffled is telling: when the original information is preserved but its sentences are reordered, accuracy stays high ($96.4$ on Qwen3-4B). The model thus relies on \emph{what} the preplan conveys rather than its surface order, confirming that the preplan carries substantive, content-level guidance. Together, these inference-time results show that PPC's adherence reward (\S\ref{sec:rl}) drives genuine, content-dependent reliance on the preplan.

\section{Conclusion}
We identified a paradigm-level gap in plan-based reasoning: while existing methods model \emph{how to solve} a problem through planning and execution, they leave \emph{what to solve}, the problem type, applicable tools, constraints, and pitfalls, entirely implicit. We proposed \textbf{PPC} (\textbf{P}replan-\textbf{P}lan-\textbf{C}oT), introducing an explicit preplan stage that gives rise to the \emph{question $\rightarrow$ preplan $\rightarrow$ plan $\rightarrow$ cot} paradigm, realized through a spoiler-score-guarded synthesis pipeline for clean preplan supervision and a composite GRPO reward that enforces the plan genuinely follows from the preplan. Extensive experiments across four backbones and five mathematical reasoning benchmarks demonstrate that PPC achieves the best results on 39 of 40 metrics, improving maj@16 and pass@16 by +2.23 and +3.06 over the strongest baseline at no additional inference-time token cost.

% while introducing no inference token overhead.

\bibliography{main}

\appendix

\section{Error Attribution Analysis}
\label{appendix:analysis}

To substantiate our claim that methods without an explicit preplan stage fail because misunderstand \emph{what} to solve rather than \emph{how}, we run an LLM-as-judge analysis over each method's incorrectly answered problems. Rather than inferring the claim indirectly, we let the judge decide it: for each wrong solution, it judges whether the failure stems from not understanding \emph{what to solve}, defined via four facets a short computation-free analysis could catch: \textit{problem type}, \textit{tools/concepts}, \textit{constraints}, and \textit{pitfalls}. The judge prompt is shown in Figure~\ref{fig:prompt}.

\begin{figure}[t]
  \centering
  \includegraphics[width=0.95\columnwidth]{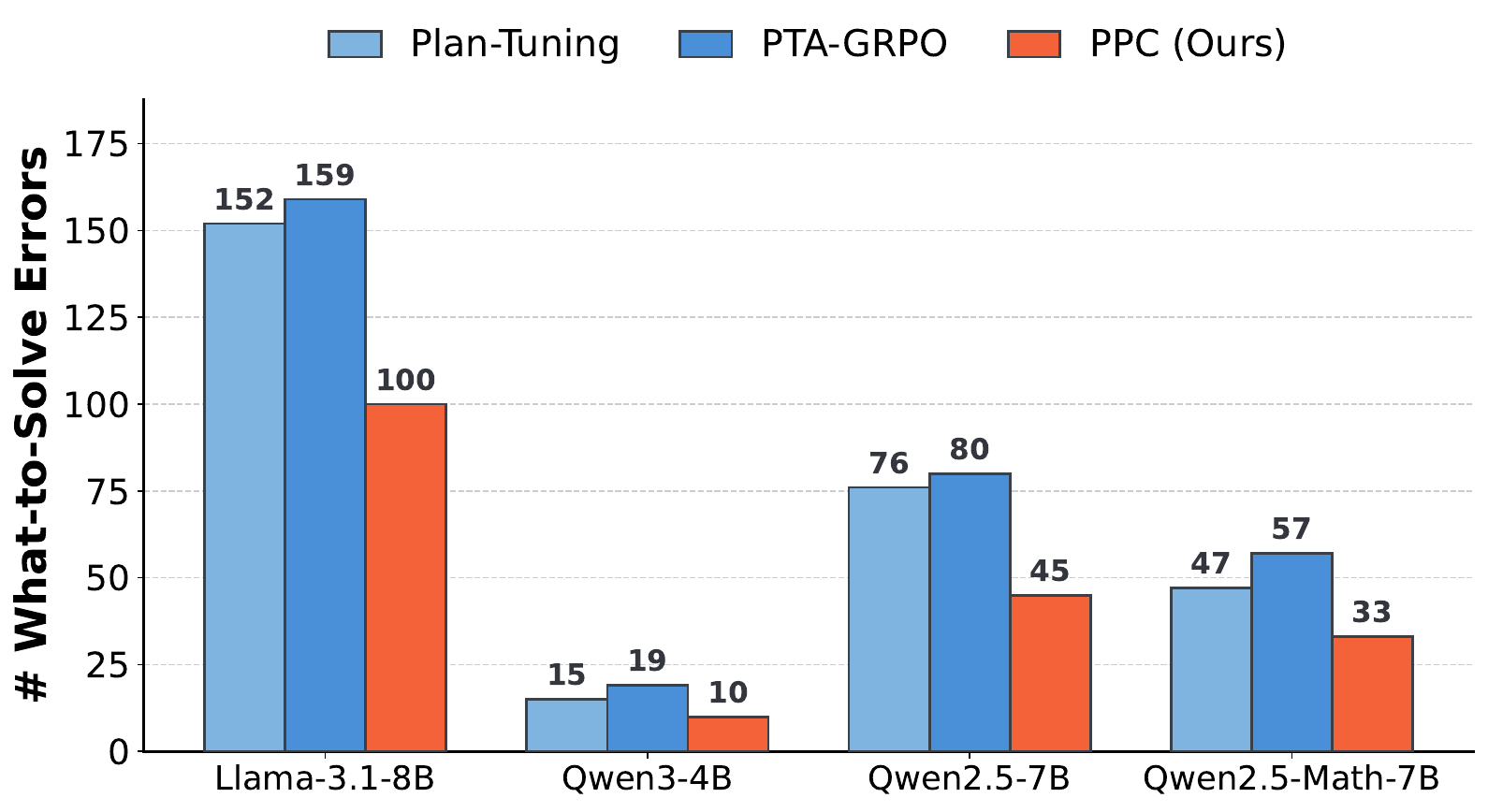}
  \caption{Number of \emph{what-to-solve} errors on MATH-500 across four backbones,
  attributed by a Qwen judge (cf. Figure~\ref{fig:motivation}, judged by DeepSeek-V4). The ranking and magnitudes closely match, indicating the attribution is robust to the choice of judge.}
  \label{fig:motivation_qwen}
  % \vskip 0.2in
\end{figure}

\noindent To verify that findings in Figure~\ref{fig:motivation} is \emph{judge-agnostic} rather than an artifact of a single judge, we repeat the identical attribution pipeline with a Qwen judge. As shown in Figure~\ref{fig:motivation_qwen}, the result is highly consistent: across all four backbones, PPC again commits substantially fewer \emph{what-to-solve} errors than the baselines, with the same ranking and comparable magnitudes. This confirms that our observation: the gains of pre-planning stem from better understanding \emph{what} to solve is robust to the choice of judge.

\begin{figure*}[t]
\begin{tcolorbox}[
    colback=gray!5,
    colframe=gray!50,
    title=Error Attribution Prompt,
    fonttitle=\bfseries,
    left=2pt, right=2pt, top=2pt, bottom=2pt,
]
\begin{lstlisting}[style=promptstyle]
You are an expert mathematician diagnosing WHY a model's solution to a competition math problem is wrong. The model produced an INCORRECT final answer.

Your single most important job is to decide ONE thing:

  >> Was this failure caused by the model not understanding WHAT TO SOLVE?

Definition you MUST use. "Understanding WHAT TO SOLVE" is the non-computational understanding of the problem that should happen BEFORE any calculation. It has exactly four facets:
  (1) PROBLEM TYPE  - recognizing what kind of problem this is and what overall approach it calls for.
  (2) TOOLS/CONCEPTS - knowing which theorem, formula, concept, or technique is the right one to bring to bear.
  (3) CONSTRAINTS   - noticing the boundary conditions, domain restrictions, edge cases, or the multiple cases that must be considered.
  (4) PITFALLS      - anticipating well-known traps (misreading a condition, double counting, sign/orientation issues baked into the setup, off-by-one in the framing).

A "WHAT-TO-SOLVE failure" means the root cause is a breakdown in one of these four facets: the model went wrong because it misjudged the nature of the problem, brought the wrong tool, ignored a constraint, or walked into a foreseeable trap -- i.e. a short upfront analysis of the problem (without doing any calculation) would plausibly have caught it.

This is the OPPOSITE of a "HOW-TO-SOLVE failure", where the model correctly understood what the problem needs and chose a sound approach, but failed while CARRYING IT OUT.

[... full label definitions, decision rules, and JSON output format omitted]
\end{lstlisting}
\end{tcolorbox}
\caption{The error attribution prompt (abridged).}
\label{fig:prompt}
\end{figure*}

\begin{figure*}[t]
\begin{preplanbox}{Discarded preplan (spoiler + leakage)}
\textit{Spoiler score $s(y_{\text{pp}})=4$; final answer correct; discarded.}\par\smallskip
This is a complex contour integral problem \dots\ using the residue theorem. \dots\
Since the contour is the circle $|z-1| = \tfrac{3}{2}$, \textbf{it encloses both
$z=0$ and $z=1$}, so we must consider the nature of each singularity and how they
contribute to the residue sum. The problem specifies using Laurent series
expansions for $e^{1/z}$ and $\tfrac{1}{z-1}$, which implies that
\textbf{understanding how to multiply and manipulate these series to extract the
coefficient of $\tfrac{1}{z}$ will be central} to determining the residue at $z=0$.
We should also be careful about distinguishing between removable singularities,
poles, and essential singularities \dots\ A key pitfall is misidentifying the order
of a pole or mishandling the series expansions, especially near the
\textbf{essential singularity at the origin}.
\end{preplanbox}

\begin{preplanbox}{Retained preplan (clean)}
\textit{Spoiler score $s(y_{\text{pp}})=2\le\tau_s$; retained.}\par\smallskip
This is a modular arithmetic problem involving large exponents and simultaneous
congruences. \dots\ The key concepts at play include the Chinese Remainder Theorem
and properties of modular exponentiation. Since $1001$ factors into the primes
$7$, $11$, and $13$, the given congruences suggest that the solution involves
combining these individual modular results into a unified modulus. A natural
approach is to treat each congruence as a piece of a larger puzzle and reconstruct
the final result modulo $1001$. We should also be careful not to assume uniqueness
without verifying compatibility of the congruences. One potential pitfall is
misapplying the theorem or mixing up moduli during calculations \dots
\end{preplanbox}
\caption{Two preplans of comparable difficulty, both reaching correct answers.
\textbf{Top:} a discarded preplan that \emph{computes and pre-enacts}—asserting computed classifications and rehearsing the solution steps. \textbf{Bottom:} a
retained preplan that \emph{analyzes and defers}—naming tools and pitfalls without solving. The spoiler score $s(\cdot)$ separates the two.}
\label{fig:preplan}
\end{figure*}

\begin{figure}[t]
\begin{promptbox}{Computation of the spoiler score $s(\cdot)$}
\begin{lstlisting}[style=promptstyle]
Input : preplan text y
Output: spoiler score s in {0,...,6}

DERIVATION_PHRASES = {"simplifies to", "reduces to",
  "leads to", "results in", "yields", "gives us",
  "implies that", "becomes", "transforms to",
  "evaluates to", "rewrites as"}
ANSWER_PHRASES = {"the answer", "the result is",
  "we get", "we obtain"}

s1 = any phrase in DERIVATION_PHRASES occurs in y
s2 = ( count("=", y) + count("\equiv", y) ) >= 3
s3 = number of inline-math spans $...$ with
        length >= 30  is >= 2
s4 = y contains a standalone integer with >= 3 digits
s5 = any phrase in ANSWER_PHRASES occurs in y
s6 = number of inline-math spans $...$  is >= 4

s = s1 + s2 + s3 + s4 + s5 + s6
return s        # retain trajectory iff s <= tau_s
\end{lstlisting}
\end{promptbox}
\caption{Rule-based computation of the spoiler score. Each of the six signals contributes one point; a preplan is retained only when $s \le \tau_s$ (we use$\tau_s = 2$).}
\label{alg:spoiler}
\end{figure}

\begin{figure*}[t]
\begin{promptbox}{Stage 1 --- Preplan Prompt}
\begin{lstlisting}[style=promptstyle]
You are a math teacher briefing a student before they attempt a problem. You yourself MUST NOT solve the problem. Your only job is to set the stage so that the student knows HOW TO APPROACH the problem mentally before starting.

<question>
{question}
</question>

Write a brief pre-solution analysis as a single coherent paragraph (around 4-8 sentences). Cover the following aspects naturally -- without using labels or section headers:
- What TYPE of problem is this? (e.g., "this is a contour integral", "this is a divisibility problem")
- What general TOOLS or CONCEPTS are likely useful? (Name them, do not apply them.)
- What is the high-level STRATEGIC DIRECTION? (Describe in plain words, not as steps.)
- What CONSTRAINTS or boundary conditions matter?
- What PITFALLS should be anticipated?

ABSOLUTE RULES (any violation invalidates your output):
1. NO derivations. Do NOT use phrases like "this simplifies to", "this leads to", "this implies", "this becomes", "we get", "the result is".
2. NO formulas longer than 15 characters. You may name a quantity (e.g., "the integral", "the polynomial f(x)") but DO NOT write extended expressions in $...$.
3. NO equations beyond definitional ones (e.g., "let n be the count of...").
4. NO specific computed values. Even if you know the answer, do NOT mention it.
5. NO step-by-step procedures. The plan comes later -- your job is meta-thinking.
6. Write as flowing prose, NOT as a list. Use natural transitions like "this suggests", "a natural approach is", "we should also be careful about".
7. Be reasonably concise -- focus on insights, not exhaustive coverage.

Output the paragraph directly. No headers, no labels, no quotation marks around the output.
\end{lstlisting}
\end{promptbox}
\caption{Stage 1 (preplan) prompt, which constrains the output to high-level analysis rather than step-by-step solving}
\label{fig:prompt_preplan}
\end{figure*}

\begin{figure*}[t]
\begin{promptbox}{Stage 2 --- Plan Prompt}
\begin{lstlisting}[style=promptstyle]
You are given a math problem and a pre-analysis written by a teacher. Your job is to translate the high-level strategic direction in the pre-analysis into a concrete numbered solution plan.

<question>
{question}
</question>

<pre_analysis>
{preplan}
</pre_analysis>

Create a numbered plan (4-7 steps). Each step should:
- Have a brief title and a one-sentence description
- Describe WHAT to do and WHY (referencing the strategy from the pre-analysis)
- NOT perform actual calculations (that comes in execution)

The plan MUST faithfully follow the strategy hinted at in the pre-analysis. Do NOT invent a different approach.

Output EXACTLY in this format (no other text):

1. [Step Title]: [one sentence describing what to do and why]
2. [Step Title]: [one sentence]
3. ...

Rules:
- Each step <= 2 sentences.
- NO specific numerical computations.
- NO LaTeX formulas longer than 20 characters.
- The total plan should be 400-1000 characters.
\end{lstlisting}
\end{promptbox}
\caption{Stage 2 (plan) prompt. The plan is conditioned on the preplan and must faithfully follow its strategy.}
\label{fig:prompt_plan}
\end{figure*}

\begin{figure*}[t]
\begin{promptbox}{Stage 3 --- Execution Cleanup Prompt}
\begin{lstlisting}[style=promptstyle]
You are given a mathematician's raw solution work for a math problem, and the solution plan that was supposed to guide it. Your task is to organize the raw solution into a clean, structured execution.

<question>
{question}
</question>

<solution_plan>
{plan}
</solution_plan>

<raw_solution>
{raw_solution}
</raw_solution>

Organize the raw solution into a structured numbered execution.
Rules:
1. Each step title MUST exactly match the corresponding plan step title.
2. Include the full mathematical reasoning and calculations from the raw solution.
3. If the raw solution contains errors, self-corrections, or multiple attempts, use the FINAL corrected version.
4. Preserve ALL numerical calculations from the raw solution. Do NOT re-derive.

Your output MUST end with exactly: Final Answer: \boxed{answer}

Output EXACTLY in this format (no other text before or after):

1. [Title matching plan step 1]: [calculation and reasoning]
2. [Title matching plan step 2]: [calculation and reasoning]
...
Final Answer: \boxed{answer}
\end{lstlisting}
\end{promptbox}
\caption{Stage 3 cleanup prompt. The raw reasoning trace is reorganized so that step titles align with the plan, preserving (not re-deriving) all computed values.}
\label{fig:prompt_cleanup}
\end{figure*}

\begin{figure*}[t]
\begin{promptbox}{Plan-adherence judge prompt ($R_{\text{adh}}$)}
\begin{lstlisting}[style=promptstyle]
You are evaluating whether a solution plan truly follows from a pre-analysis (preplan).

Question: {question}
Pre-analysis (preplan): {preplan}
Solution plan: {plan}

Rate how well the plan FOLLOWS the preplan's strategic direction (1-5):
- 5: Plan tightly implements the strategy hinted in preplan; the tools/concepts mentioned in preplan appear in the plan steps; the plan would be DIFFERENT if the preplan suggested a different approach.
- 4: Mostly follows, with minor unrelated additions or one missing element.
- 3: Partially aligns; some strategic elements reflected, but the plan also wanders into directions the preplan did not hint at.
- 2: Only loosely connects; the plan would look largely the same even under a different preplan.
- 1: Completely ignores or contradicts the preplan.

IMPORTANT: Score by STRATEGY ALIGNMENT, not by quality. A correct plan that ignores the preplan should still score LOW; an OK plan that faithfully follows it should score HIGH.
Output ONLY a single integer 1-5. No explanation.
\end{lstlisting}
\end{promptbox}
\caption{The frozen plan-adherence judge prompt for $R_{\text{adh}}$.}
\label{fig:prompt_adh}
\end{figure*}

\section{Spoiler/Leakage Failure: A Full Example} \label{app:spoiler_example} 
This appendix expands on the spoiler-score filter introduced in \S\ref{sec:sft}. Figure~\ref{fig:preplan} contrasts a \emph{discarded} preplan that crosses the conceptual integrity boundary with a \emph{retained} clean preplan on a problem of comparable difficulty. Both trajectories reach a correct final answer; only the first is discarded, showing that the filter targets preplan purity, not answer correctness. 

\paragraph{Why the first is discarded.} The discarded preplan does not survey \emph{what to solve}; it has already settled \emph{how to solve}. It commits to concrete computed outcomes that both singularities are enclosed and that the origin is an \emph{essential} singularity (a spoiler), and rehearses the exact downstream procedure of expanding both Laurent series, multiplying, and extracting the $1/z$ coefficient (a leakage). These trip the derivation-phrasing and answer-revealing signals of $s(\cdot)$, pushing its score above $\tau_s$ despite the correct answer $2\pi i$. 

\paragraph{Why the second is retained.} The retained preplan stays on the \emph{what-to-solve} side throughout: it names the relevant tool (CRT), notes the structural fact $1001 = 7\cdot 11\cdot 13$, and flags genuine pitfalls—without solving any congruence or committing to a step sequence. The actual reconstruction is left to the planning and execution stages.

\section{Three-Stage SFT Data Generation Prompts}
\label{app:three_stage_prompts}

This appendix provides the full prompts used to construct the three-stage $\langle$preplan, plan, execute$\rangle$ trajectories of \S\ref{sec:sft}. The stages are generated strictly left-to-right, and each stage's prompt is shown below (with the runtime fields \texttt{\{question\}}, \texttt{\{preplan\}}, \texttt{\{plan\}} substituted at generation time).

\paragraph{Stage 1: Preplan.}
The preplan is produced first, before any plan exist. The prompt (Figure~\ref{fig:prompt_preplan}) casts the planner LLM as a teacher \emph{briefing} a student rather than solving the problem, and enforces the conceptual integrity through explicit rules: no derivations, no extended formulas, no computed values, and prose rather than steps. These rules operationalize the spoiler constraint that the filter of \S\ref{sec:sft} later enforces.

\paragraph{Stage 2: Plan.}
Conditioned on the question \emph{and} the preplan, the same instruction-tuned LLM expands the high-level direction into a concrete numbered plan (Figure~\ref{fig:prompt_plan}). The prompt requires the plan to faithfully follow the strategy hinted at in the preplan, which is what makes the plan a genuine elaboration of the preplan rather than an independent solution.

\paragraph{Stage 3: Execute.}
The executor sees only the question and the plan, never the preplan, so that execution is driven by the explicit plan alone. We realize execution with a strong reasoning model (DeepSeek-R1) that produces a raw worked solution, which is then reorganized by the instruction-tuned model into a clean numbered execution whose step titles align one-to-one with the plan (Figure~\ref{fig:prompt_cleanup}). The cleanup step preserves all numerical results from the raw solution rather than re-deriving them, keeping the mathematics intact while imposing a consistent structure.

\paragraph{Verification.}
A trajectory is kept only if its final answer matches the gold answer (verified by normalized string match with an LLM equivalence check as fallback) and its execution ends in the required \texttt{\textbackslash boxed\{\}} format.

\section{The Spoiler Score \texorpdfstring{$s(\cdot)$}{s}}
\label{app:spoiler_score}

This appendix specifies the rule-based spoiler score $s(\cdot)$ used by the filter in Eq.~\ref{eq:filter}. The score is an integer in $\{0,\dots,6\}$ obtained by summing six binary signals, each of which fires when the preplan shows a symptom of \emph{computing} or \emph{revealing the answer} rather than \emph{analyzing} the problem. A preplan is retained only when $s(y_{\text{pp}}) \le \tau_s$; we use $\tau_s = 2$. The score is applied uniformly at construction time (Eq.~\ref{eq:filter}) and as a reward component during RL without inconsistency. 

\paragraph{The six signals.} 
Let $y_{\text{pp}}$ be the candidate preplan text. Each signal below contributes $1$ to $s(y_{\text{pp}})$ when its condition holds: \begin{enumerate} 
\item \textbf{Derivation phrasing.} The text contains any phrase signalling an inference or transformation has been carried out: e.g.\ ``simplifies to'', ``reduces to'', ``leads to'', ``yields'', ``implies that'', ``becomes''. Such phrases indicate the preplan is performing, not previewing, the derivation. 
\item \textbf{Equation density.} The combined count of equality symbols (\texttt{=} and \texttt{\textbackslash equiv}) is at least three. A preplan that states constraints needs few equalities; a high count signals worked relations. 
\item \textbf{Long inline expressions.} At least two inline-math spans exceed $30$ characters. Short named quantities are permitted, but extended expressions indicate concrete manipulation. 
\item \textbf{Multi-digit constants.} The text contains a standalone integer of three or more digits, which typically appears only once specific numbers are being computed or stated. 
\item \textbf{Answer-revealing wording.} The text contains phrases such as ``the answer'', ``the result is'', ``we get'', or ``we obtain'', which tend to precede a disclosed outcome. 
\item \textbf{Inline-math span count.} The total number of inline-math spans is at least four, indicating the preplan leans on formal expressions rather than prose analysis. \end{enumerate} 
\paragraph{Aggregation and rationale.} The final score sums the six indicators, 

\begin{equation} 
s(y_{\text{pp}}) \;=\; \sum_{i=1}^{6} {1}\!\left[\,\text{signal}_i \text{ fires on } y_{\text{pp}}\,\right] \;\in\; \{0,\dots,6\}, 
\end{equation} 
This additive design tolerates the incidental use of light notation that even a purely analytical preplan may contain, and targets the \emph{accumulation} of computational behaviour that characterizes a true spoiler. The procedure is summarized in Figure~\ref{alg:spoiler}.

\section{Reward Details}
\label{app:reward_details}
In this section, we show the details of each reward components. All judges are \emph{frozen} (a fixed instruction-tuned LLM, not updated during RL), so the reward signal is stationary across training.

\paragraph{Outcome ($R_{\text{out}}$).}
To provide a gradient beyond a binary signal, so that GRPO can still learn once most rollouts are wrong, an incorrect answer is scored by a \emph{proximity} judge on a $1$--$5$ scale (mapped to $[0-0.5]$) that rates how close the \emph{solution path} was to a correct one, deliberately ignoring numerical closeness of the final value.

\paragraph{Format ($R_{\text{fmt}}$).}
A rule-based check awards $0.3$ iff the completion has a clean structure: the three tags \texttt{<preplan>}, \texttt{<plan>}, \texttt{<execute>} each appear exactly once, in order, with a \texttt{\textbackslash boxed\{\}} answer inside \texttt{<execute>}, no trailing text after \texttt{</execute>}, and no tags outside the allowed set.

\paragraph{Style ($R_{\text{sty}}$).}
To keep the preplan from regressing into a spoiler, we reuse the spoiler score $s(\cdot)$ of Appendix~\ref{app:spoiler_score} as a one-sided penalty $R_{\text{sty}} = \max(0,\, s(y_{\text{pp}}) - \tau_s)$ with $\tau_s = 2$, weighted by $\lambda_s = 0.1$ in Eq.~\eqref{eq:reward}. Thus the penalty term $-\lambda_s R_{\text{sty}}$ is $0$ while $s \le 2$ and decreases (down to $-0.4$ when $s$ reaches its maximum of $6$) as the preplan accumulates computational signals.

\paragraph{Adherence ($R_{\text{adh}}$).}
A frozen judge rates, on a $1$--$5$ scale, whether the plan implements the \emph{strategy} of the preplan (not its quality), scaled to a maximum of $0.1$. Crucially, a plan that solves the problem well but ignores the preplan still scores low, which is what ties the plan to the preplan rather than letting it become an independent solution. The full prompt is shown in Figure~\ref{fig:prompt_adh}.

% ───────────────── Stage 3b: Cleanup ─────────────────

\end{document}